\documentclass[letterpaper, 10 pt, conference]{ieeeconf}  % Comment this line out if you need a4paper

\IEEEoverridecommandlockouts                              % This command is only needed if 
\usepackage{aircraftshapes}
\usepackage{vector}
\usepackage{tikz}
\usepackage{pgfplots}
\pgfplotsset{compat=newest}
\pgfplotsset{every axis legend/.append style={%
		cells={anchor=west}}
}
\pgfplotsset{every y tick label/.append style={font=\footnotesize}}
\pgfplotsset{every x tick label/.append style={font=\footnotesize}}
\pgfplotsset{every axis x label/.append style={font=\footnotesize}}
\pgfplotsset{every axis y label/.append style={font=\footnotesize}}
\pgfplotsset{every axis legend/.append style={font=\footnotesize}}
\pgfplotsset{every axis title/.append style={font=\footnotesize}}

\usepgfplotslibrary{polar}
\usepgfplotslibrary{groupplots}
\usepackage[per-mode=symbol,detect-all]{siunitx}

\pgfplotscreateplotcyclelist{my colors}{
	black!50\\
	green\\
	red\\
	blue\\
}
\pgfplotsset{
	cycle list name=my colors,
	legend cell align=left,
}

\tikzset{
	compass/.pic = {
		\filldraw[pic actions,rotate=0,scale=.7,ultra thin] (0,0) -- (45:0.15)--(0:0.45) node[transform shape,rotate=0,right]{E};
		\filldraw[pic actions,fill=white,rotate=0,scale=.7,very thin] (0,0) -- (-45:0.15)--(0:0.425)--cycle;

		\filldraw[pic actions,rotate=90,scale=.7,ultra thin] (0,0) -- (45:0.15)--(0:0.45) node[transform shape,rotate=-90,above]{N};
		\filldraw[pic actions,fill=white,rotate=90,scale=.7,very thin] (0,0) -- (-45:0.15)--(0:0.425)--cycle;

		\filldraw[pic actions,rotate=180,scale=.7,ultra thin] (0,0) -- (45:0.15)--(0:0.45) node[transform shape,rotate=-180,left]{W};
		\filldraw[pic actions,fill=white,rotate=180,scale=.7,very thin] (0,0) -- (-45:0.15)--(0:0.425)--cycle;

		\filldraw[pic actions,rotate=270,scale=.7,ultra thin] (0,0) -- (45:0.15)--(0:0.45) node[transform shape,rotate=-270,below]{S};
		\filldraw[pic actions,fill=white,rotate=270,scale=.7,very thin] (0,0) -- (-45:0.15)--(0:0.425)--cycle;
	}
}

\DeclareGraphicsExtensions{.png}
\usepackage{amssymb}
\usepackage{amsmath}
\usepackage{dsfont}
\usepackage[T1]{fontenc}
\usepackage[ansinew]{inputenc}
\usepackage{array}
\usepackage{fixltx2e}
\usepackage{float}
\usepackage{url}
\usepackage{booktabs}
\usepackage[capitalize]{cleveref}

\usepackage[backend=bibtex,style=ieee]{biblatex}
\makeatletter
\def\blx@maxline{77}
\makeatother

\makeatletter
\def\BState{\State\hskip-\ALG@thistlm}
\makeatother

\AtEveryBibitem{
	\ifentrytype{inproceedings}{
		\clearlist{address}
		\clearlist{publisher}
		\clearname{editor}
		\clearlist{organization}
		\clearfield{url}  
		\clearfield{url}  
		\clearfield{doi}  
		\clearfield{pages}  
		\clearlist{location}}{}}
\AtEveryBibitem{
	\clearfield{doi}
	\clearfield{issn}
	\clearfield{month}  
}
\AtEveryBibitem{
	\ifentrytype{article}{
		\clearlist{address}
		\clearlist{publisher}
		\clearname{editor}
		\clearlist{organization}
		\clearfield{url}  
		\clearfield{doi}  
		\clearlist{location}}{}}

\title{\LARGE \bf
	Image-based Guidance of Autonomous Aircraft for Wildfire Surveillance and Prediction
}

\author{Kyle D. Julian$^{1}$ and Mykel J. Kochenderfer$^{2}$% <-this % stops a space
	\thanks{$^{1}$Kyle D. Julian is a Ph.D. candidate in the Department of Aeronautics and Astronautics,
		Stanford University, Stanford, CA, 94305
		{\tt\small kjulian3@stanford.edu}}%
	\thanks{$^{2}$Mykel J. Kochenderfer is an Associate Professor in the Department of Aeronautics and Astronautics,
	Stanford University, Stanford, CA, 94305
	{\tt\small mykel@stanford.edu}}%
}

\begin{document}
	\maketitle
	\thispagestyle{empty}
	\pagestyle{empty}

\begin{abstract}
	Small unmanned aircraft can help firefighters combat wildfires by providing real-time surveillance of the growing fires. However, guiding the aircraft autonomously given only wildfire images is a challenging problem. This work models noisy images obtained from on-board cameras and proposes two approaches to filtering the wildfire images. The first approach uses a simple Kalman filter to reduce noise and update a belief map in observed areas. The second approach uses a particle filter to predict wildfire growth and uses observations to estimate uncertainties relating to wildfire expansion. The belief maps are used to train a deep reinforcement learning controller, which learns a policy to navigate the aircraft to survey the wildfire while avoiding flight directly over the fire. Simulation results show that the proposed controllers precisely guide the aircraft and accurately estimate wildfire growth, and a study of observation noise demonstrates the robustness of the particle filter approach.
\end{abstract}

\section{Introduction}

Over the past three decades, the frequency and severity of wildfires in North America has grown \cite{Schoennagel4582}. From 2006 to 2015, the US Congress spent \$13 billion to suppress wildfires \cite{hoover2015wildfire}, and a record high 10.1 million acres of land were consumed by wildfires in 2015 \cite{FireCenter}. Modeling and tracking wildfire growth allows firefighters to more effectively mitigate wildfires by distributing fire suppressant. Current approaches use computer models to predict wildfire growth \cite{andrews2007predicting} or satellite images to track wildfires \cite{martin1999fire}, but neither approach offers real-time high-resolution maps of wildfires. Because wildfires can change speed and direction due to changes in wind and fuel conditions, better accuracy is needed for firefighters to monitor wildfires.

Unmanned aerial vehicles (UAVs) are increasingly popular for wildfire surveillance because they offer real-time observations without risking human pilots \cite{ambrosia2015selection,watts2012unmanned}. Julian and Kochenderfer proposed a method for wildfire monitoring with a team of small autonomous UAVs, which was shown scale with larger numbers of aircraft \cite{julian2018autonomous}. To enable automatic control of an aircraft, images of a wildfire must be processed and used to generate trajectories that maneuver the aircraft to track moving fires. Flight tests with autonomous helicopters over controlled fires demonstrated wildfire tracking through image feature matching \cite{merino2012unmanned}. Other works developed algorithms for detecting fire within images through genetic fuzzy logic \cite{kukreti2016detection} and color space conversion with threshold segmentation \cite{yuan2015uav}. This work does not model image processing but assumes wildfire locations can be extracted from raw images.

Given noisy images of wildfire locations, two approaches are proposed to filter and compile the observations into a wildfire map, referred herein as a belief map. The first approach is based on the Kalman filter \cite{kalman1960}, which was originally developed for filtering noisy observations and has since been applied to many image-based applications, such as restoring noisy images \cite{biemond1983fast} and predicting the time evolution of roadway scenes based on images of traffic \cite{dellaert1997robust}. This work performs image-based belief updates using an extended Kalman filter (EKF), which revises a wildfire belief map in locations where observations are made.

The second proposed filtering approach uses a particle filter to simulate and estimate the extent of wildfires given observations. Particle filters are popular for filtering and prediction for non-linear systems \cite{arulampalam2002tutorial} and have been used to track moving cars \cite{niknejad2012road} and people \cite{breitenstein2009robust}. In addition, particle filters have been used with observations from multiple aircraft in order to track ground targets \cite{ong2006decentralised}. This work implements a particle filter to create an accurate wildfire belief map and predict wind parameters governing the wildfire expansion, which would give firefighters a more accurate prediction of the wildfire's location and future growth.

After creating wildfire belief maps, a deep reinforcement learning (DRL) controller is trained to navigate the aircraft. DRL trains a policy mapping high dimensional or image inputs to control actions, allowing images to be used as the basis for control. DRL controllers are trained through simulation, resulting in intelligent controllers for many applications such as Atari games and Go \cite{mnih2015human,silver2016mastering}. In previous work, DRL was used to train a neural network controller that makes decisions based on observations of wildfire \cite{julian2018autonomous}. This work considers a team of two aircraft, though more aircraft can be incorporated by considering pairwise interactions~\cite{julian2018autonomous}. This work extends the previous approach by modeling camera images, incorporating observation errors, filtering observations into a belief map, and predicting wildfire growth through wind estimation. Furthermore, this work explores how flight directly over fires, where air could be turbulent and dangerous, can be discouraged. Simulated experiments show that DRL intelligently guides aircraft to survey wildfires while accurately estimating wildfire wind and future growth.

\section{Problem Formulation}
This section describes how the wildfire, aircraft, and cameras are modeled.

\subsection{Wildfire Model}\label{sec:wildfire}
A stochastic wildfire model~\cite{Bertsimas2017} is used to simulate wildfires, as done in previous work~\cite{julian2018autonomous}. An area of land is divided into a $100\times 100$ grid of cells, and each cell $s$ has some amount of fuel $F(s)$ and is either burning ($B(s)=1$), or not burning ($B(s)=0$). A non-burning cell might begin burning at the next time step if nearby cells are burning, and the probability of ignition increases with proximity to burning cells. The fuel decreases as the cell burns until the fuel is gone and the cell extinguishes. Assuming a burning rate $\beta=1$, the wildfire propagates according to
\begin{equation}
F_{t+1}(s) = 
\begin{cases}
\max(0,F_t(s)-\beta)& \text{if }  B_t(s)\\
F_t(s)              & \text{otherwise}
\end{cases}
\end{equation}	
\begin{equation} \label{eq:probIgnite}
p(s) = 
\begin{cases}
1 -\prod_{s'} (1-P(s,s')B_t(s')) & \text{if } F_t(s)>0 \\
0                         & \text{otherwise}
\end{cases}
\end{equation} 
where $p(s)$ is the probability cell $s$ will ignite, and $P(s,s')$ is the probability cell $s'$ ignites cell $s$. Typically, $P(s,s')$ is smaller when the two cells are further apart. Wind can be modeled by biasing $P(s,s')$ in a given direction. 

\begin{figure}
	\centering
	\vspace{1pt}
	\begin{tikzpicture}[]
\begin{groupplot}[group style={horizontal sep=0.3cm, vertical sep=0.3cm, group size=2 by 2}]
\nextgroupplot [ylabel = {$T = \SI{0}{\second}$}, width = {4.8cm}, title = {Fire Locations}, height = {4.8cm}, ticks=none,yticklabels={,,}, enlargelimits = false, axis on top]\addplot [point meta min=0, point meta max=1] graphics [xmin=10, xmax=1000, ymin=10, ymax=1000] {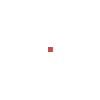};
\nextgroupplot [width = {4.8cm}, title = {Fuel Map}, height = {4.8cm}, ticks=none,yticklabels={,,}, enlargelimits = false, axis on top]\addplot [point meta min=0, point meta max=20] graphics [xmin=10, xmax=1000, ymin=10, ymax=1000] {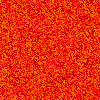};

%\nextgroupplot [ylabel = {$T = \SI{100}{\second}$}, width = {4.8cm}, height = {4.8cm}, ticks=none,yticklabels={,,}, enlargelimits = false, axis on top]\addplot [point meta min=0, point meta max=1] graphics [xmin=10, xmax=1000, ymin=10, ymax=1000] {FireMapWindy_T40.png};
%\nextgroupplot [width = {4.8cm}, height = {4.8cm}, ticks=none,yticklabels={,,},  enlargelimits = false, axis on top]\addplot [point meta min=0, point meta max=20] graphics [xmin=10, xmax=1000, ymin=10, ymax=1000] {FuelMapWindy_T40.png};

\nextgroupplot [ylabel = {$T = \SI{150}{\second}$}, width = {4.8cm},height = {4.8cm}, ticks=none,yticklabels={,,}, enlargelimits = false, axis on top]\addplot [point meta min=0, point meta max=1] graphics [xmin=10, xmax=1000, ymin=10, ymax=1000] {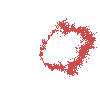};

\nextgroupplot [width = {4.8cm}, height = {4.8cm}, ticks=none,yticklabels={,,}, enlargelimits = false, axis on top]\addplot [point meta min=0, point meta max=20] graphics [xmin=10, xmax=1000, ymin=10, ymax=1000] {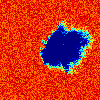};

\end{groupplot}

\pic[fill=black,text=black] at (0.75,0.75) {compass};

\end{tikzpicture}
	\caption{Wildfire propagation over time}
	\label{fig:FireProp}
\end{figure}
\Cref{fig:FireProp} shows a wildfire spreading from an initial seed with wind blowing to the east, making the fire grow more quickly towards the east. The wildfire locations are updated once every \SI{2.5}{\second}. This simplistic wildfire model illustrates the proposed algorithms, but expert knowledge of wildfires can be incorporated by changing the growth probabilities or by adding new terms to the model. Any wildfire model that is easily simulated can be used with the presented approach.

\subsection{Aircraft Model}
Autonomous aircraft are tasked with monitoring wildfire growth. Assuming the aircraft fly at constant altitude and constant speed $v$, the position $(x,y)$ and heading direction $\psi$ of the aircraft change according to 
\begin{equation}
\dot{x} = v\cos(\psi), \quad \dot{y} = v \sin(\psi), \quad \dot{\psi} = \frac{g \tan(\phi)}{v}
\end{equation}
where $g$ is the sea-level gravitational acceleration constant and $\phi$ is the aircraft bank angle. The aircraft trajectory can be controlled through $\phi$. In this work, $v=\SI{20}{\meter\per\second}$, and wind is not incorporated into the aircraft dynamics, although wind terms could be added to $\dot{x}$ and $\dot{y}$ to model the effect of wind.

\subsection{Camera Model}
Unlike prior work that assumed a fixed sensing radius around each aircraft~\cite{julian2018autonomous}, this work models fixed cameras on the aircraft to generate wildfire observations. To define the camera orientation, the aircraft orientation must first be defined. The aircraft frame is rotated with respect to the fixed north-east-down world frame using the roll-pitch-yaw convention. The aircraft body frame is centered at the aircraft center of gravity such that the $x$-axis points to the aircraft nose, the $z$-axis points downward, and the $y$-axis points out the right wing of the aircraft. In this work, the aircraft is assumed to remain at steady-level flight with pitch angle $\theta=0^\circ$. Conversion from the body frame to the world frame is accomplished by rotating about the $z$-axis by heading angle $\psi$ and then rotating about the $x$-axis by bank angle $\phi$. 

The camera frame is attached to the camera with the $x$-axis and $y$-axis defining the image plane and the $z$-axis perpendicular to the image plane and positive in front of the camera. The camera is rotated with respect to the aircraft body axes. Let $\phi_c$ be the camera rotation about the aircraft $x$-axis and $\theta_c$ be the subsequent camera rotation about the $y$-axis. As a result, a point $\vect{p}_c$ defined in the camera frame has world frame coordinates $\vect{p}_w$ where
\begin{equation}
\vect{p}_w =\vect{R}_z(\theta)\vect{R}_x(\phi)\vect{R}_y(\theta_c)\vect{R}_x(\phi_c)\vect{p}_c.
\end{equation}

A simple pinhole camera model was used, which uses a camera matrix based on focal length and assumes no distortion or skew \cite{trucco1998introductory}. With camera image coordinates $(u,v)$, focal length $f$, and world points in the camera frame $(x,y,z)$, the pinhole camera model defines
\begin{equation}\label{eq:cam_mat}
(u,v)=(\frac{fx}{z},\frac{fy}{z}).
\end{equation}

Because \cref{eq:cam_mat} reduces the dimensionality of observations from 3-D to 2-D, there are infinitely many $(x,y,z)$ points that give the same $(u,v)$. However, the aircraft is assumed to be flying at altitude $h=\SI{200}{\meter}$, so there will only be one ground point for every image point. The cameras were modeled as \SI{35}{\milli\meter} cameras with \SI{24}{\milli\meter} by \SI{36}{\milli\meter} images and \SI{50}{\milli\meter} focal lengths. In this work, the aircraft use four cameras with $\theta_c=30^\circ$ and $\phi_c\in [-40^\circ,-13^\circ,13^\circ,40^\circ]$. In addition, sensor observations are limited to a range of \SI{300}{\meter} to prevent the aircraft from viewing points very far away. 

\begin{figure}
    \vspace{4pt}
	\begin{tikzpicture}[]
\begin{groupplot}[group style={horizontal sep=1.0cm, group size=2 by 1}]
\nextgroupplot [height = {5cm}, ylabel = {North}, xmin = {0}, xmax = {1000}, ymax = {1000}, xlabel = {East}, ticks=none, ymin = {0}, width = {5cm}]\addplot+ [no marks, ultra thick, solid, gray]coordinates {
(700, 620)
(720, 560)
(770, 570)
(850, 590)
(850, 600)
(830, 650)
(820, 670)
(800, 670)
(730, 640)
(710, 630)
(700, 620)
};
\node[] at (200,900) {\footnotesize$\phi=0^\circ$};
\node[] at (700,750) {\footnotesize$\phi=40^\circ$};
\addplot+ [no marks, ultra thick, solid, gray]coordinates {
(840, 590)
(720, 560)
(730, 510)
(740, 470)
(930, 470)
(920, 490)
(860, 580)
(850, 590)
(840, 590)
};
\addplot+ [no marks, ultra thick, solid, gray]coordinates {
(990, 140)
(990, 400)
(970, 430)
(920, 480)
(810, 480)
(740, 470)
(740, 380)
(750, 270)
(800, 240)
(930, 170)
(990, 140)
};
\addplot+ [no marks, ultra thick, solid, gray]coordinates {
(740, 0)
(990, 0)
(990, 160)
(980, 170)
(800, 260)
(750, 280)
(740, 30)
(740, 0)
};
\addplot+ [no marks, ultra thick, solid, gray]coordinates {
(200, 760)
(220, 680)
(270, 680)
(390, 710)
(470, 950)
(470, 960)
(460, 960)
(330, 870)
(230, 800)
(200, 770)
(200, 760)
};
\addplot+ [no marks, ultra thick, solid, gray]coordinates {
(230, 600)
(370, 600)
(380, 620)
(390, 700)
(390, 710)
(370, 710)
(330, 700)
(260, 680)
(230, 670)
(230, 600)
};
\addplot+ [no marks, ultra thick, solid, gray]coordinates {
(230, 600)
(230, 530)
(260, 520)
(330, 500)
(370, 490)
(390, 490)
(390, 500)
(380, 580)
(370, 600)
(230, 600)
};
\addplot+ [no marks, ultra thick, solid, gray]coordinates {
(200, 430)
(230, 400)
(330, 330)
(460, 240)
(470, 240)
(470, 250)
(390, 490)
(270, 520)
(220, 520)
(200, 440)
(200, 430)
};
\node[aircraft top,fill=black,draw=white, minimum width=2.0cm,rotate=0.0,scale = 0.30] at (axis cs:200.01452236049005, 600.0026932343611) {};;
\node[aircraft top,fill=cyan,draw=white, minimum width=2.0cm,rotate=0.0,scale = 0.30] at (axis cs:700.01452236049, 600.0026932343611) {};;
\nextgroupplot [height = {5cm}, ylabel = {North}, xmin = {0}, xmax = {1000}, ymax = {1000}, xlabel = {East}, ticks=none, ymin = {0}, width = {5cm}]\addplot+ [no marks, ultra thick, solid, gray]coordinates {
(700, 620)
(720, 560)
(770, 570)
(850, 590)
(850, 600)
(830, 650)
(820, 670)
(800, 670)
(730, 640)
(710, 630)
(700, 620)
};
\node[] at (200,900) {\footnotesize$\phi=0^\circ$};
\node[] at (700,750) {\footnotesize$\phi=40^\circ$};
\addplot+ [no marks, ultra thick, solid, gray]coordinates {
(840, 590)
(720, 560)
(730, 510)
(740, 470)
(930, 470)
(920, 490)
(860, 580)
(850, 590)
(840, 590)
};
\addplot+ [no marks, ultra thick, solid, gray]coordinates {
(740, 470)
(740, 380)
(750, 310)
(770, 310)
(830, 330)
(880, 360)
(940, 420)
(950, 440)
(950, 450)
(920, 480)
(810, 480)
(740, 470)
};
\addplot+ [no marks, ultra thick, solid, gray]coordinates {
(330, 870)
(230, 800)
(200, 770)
(200, 760)
(220, 680)
(270, 680)
(390, 710)
(420, 800)
(380, 840)
(330, 870)
};
\addplot+ [no marks, ultra thick, solid, gray]coordinates {
(230, 600)
(370, 600)
(380, 620)
(390, 700)
(390, 710)
(370, 710)
(330, 700)
(260, 680)
(230, 670)
(230, 600)
};
\addplot+ [no marks, ultra thick, solid, gray]coordinates {
(230, 600)
(230, 530)
(260, 520)
(330, 500)
(370, 490)
(390, 490)
(390, 500)
(380, 580)
(370, 600)
(230, 600)
};
\addplot+ [no marks, ultra thick, solid, gray]coordinates {
(200, 440)
(200, 430)
(230, 400)
(330, 330)
(380, 360)
(420, 400)
(390, 490)
(270, 520)
(220, 520)
(200, 440)
};
\node[aircraft top,fill=black,draw=white, minimum width=2.0cm,rotate=0.0,scale = 0.30] at (axis cs:200.01452236049005, 600.0026932343611) {};;
\node[aircraft top,fill=cyan,draw=white, minimum width=2.0cm,rotate=0.0,scale = 0.30] at (axis cs:700.01452236049, 600.0026932343611) {};;
\end{groupplot}

\end{tikzpicture}
	\vspace{-15pt} 
	\caption{Aircraft imaging regions without limiting range (left) and with a range limit of 300 meters (right)}
	\label{fig:Sensors}
\end{figure}
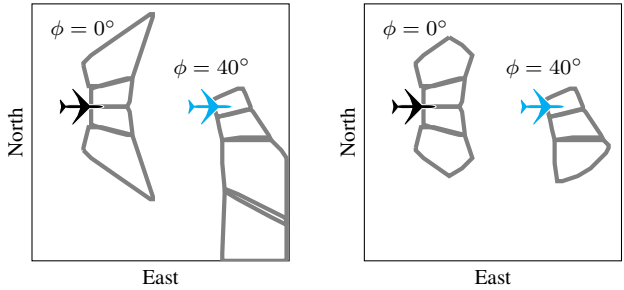

\Cref{fig:Sensors} shows the observations for an aircraft in level flight and one banked $40^\circ$ during a left turn. Each aircraft observes four regions, one for each camera. The bank angle significantly skews the observations because the aircraft bank angle also rotates the attached cameras. The range of the images is limited to \SI{300}{\meter}, as shown in the right plot of \cref{fig:Sensors}, so the aircraft are limited to viewing only nearby locations.

Each observation image maps a region of 30 by 20 pixels from image coordinates to the ground where the wildfire of the nearest cell is observed. To introduce observation errors, 10\% of the image observations are changed at random. Noisy images are filtered and compiled into a belief map that specifies all locations believed to be burning. By compiling observations into a belief map, information over the course of many observations can be preserved and used to make informed decisions. Two methods for filtering images are presented in \cref{sec:EKF,sec:PF}.

\section{Deep Reinforcement Learning Controller}
	A controller is needed to command bank angle $\phi$ so that aircraft cooperatively monitor a wildfire. One method to map state information to control commands is to train a neural network controller through deep Q-learning, a deep reinforcement learning (DRL) algorithm~\cite{mnih2015human}. Deep Q-learning trains a neural network through simulation to represent the state-action value function $Q(s,a)$. When considering two aircraft, the state information for one aircraft is composed of the relative pose of the other aircraft and wildfire locations as follows:
	\begin{enumerate}
		\item $\rho$ (\si{\meter}): Distance to the other aircraft
		\item $\theta_r$ (\si{\radian}): Angle to other aircraft relative to the aircraft's heading direction
		\item $\psi_r$ (\si{\radian}): Heading angle of other aircraft relative to the aircraft's heading direction
		\item $\phi_0$ (\si{\radian}): Bank angle of aircraft
		\item $\phi_1$ (\si{\radian}): Bank angle of other aircraft
		\item $\text{Belief}_r$ (image): Wildfire belief map relative to aircraft position and heading
	\end{enumerate}
	Therefore, the state information differs for each aircraft and describes the scenario from each aircraft's perspective.
	
	In this work, there are two possible actions: to increase or decrease commanded bank angle by $\SI{5}{\deg}$, which could be tracked using a proportional-derivative controller. Bank angle commands are given at a frequency of $\SI{10}{\hertz}$ and limited to $\phi \in [\SI{-50}{\deg},\SI{50}{\deg}]$, which allows the controller to quickly change the commanded bank angle while still commanding precise angles.
	
	In deep Q-learning, simulations are conducted to collect tuples of state $s$, action $a$, reward $r$, and next state $s'$. For this application, the reward function is designed to encourage desired behavior and includes
	\begin{enumerate}
		\item Reward for new wildfire locations in belief map
		\item Penalty for proximity to other aircraft
		\item Penalty for flying directly over wildfire
	\end{enumerate}
	Rewarding the aircraft for new wildfire locations appearing in the belief map encourages the aircraft to observe areas where the wildfire is expanding. Penalizing the aircraft for flying near the other aircraft encourages the aircraft to make distinct observations, and penalizing the aircraft for flying over the wildfire discourages the aircraft from flying in potentially turbulent and dangerous conditions. The effect of the last penalty is explored in greater detail in \cref{sec:results}.
	
	To approximate the state-action values, $Q(s,a)$ is parameterized by a neural network. For this application, the neural network uses ReLU activations and is composed of fully connected layers for the five continuous state variables and convolutional layers with max-pooling for the relative belief map image. The flattened output of the convolutional layers is concatenated with the output of the fully connected layers and followed by two more fully connected layers. The output layer is composed of two values, representing the value of the two possible actions. 
	
	For a tuple $(s,a,r,s')$, state-action values ideally follow
	\begin{equation}\label{eq:qlearning}
	Q(s,a) = r + \gamma \max_{a'} Q(s',a') 
	\end{equation}
	where $\gamma$ is a discount factor, set to $\gamma=0.99$ here. Gradient descent methods are used to update the neural network parameters to minimize errors in \cref{eq:qlearning}, as described by~\citeauthor{mnih2015human}~\cite{mnih2015human}.

\section{Extended Kalman Filter}\label{sec:EKF}
One approach to filtering noisy observations uses an extended Kalman filter (EKF). Let $\vect{x}_t \in \mathbb{R}^N$ be the current state where $N$ is the number of cells in the grid and $x_t^i$ represents the probability that cell $i$ is burning at time $t$. The EKF estimates $\vect{x}_t$ with a mean value $\vect{mu}_t$ and covariance $\vect{\Sigma}_t$. With the assumption that the fire spreads slowly, $\vect{x}_t$ can be approximated as static subject to Gaussian white process noise, which allows the Jacobian of the state dynamics to be reduced to an identity matrix. Furthermore, the observation made at each point in the observation image depends on the burning status of the cells closest to the point of observation, so the observation Jacobian for a given point in the observation is 1 for cells closest to that point and 0 for other cells.

However, the EKF approach has a significant drawback. With $N=10000$, the matrix inversion of $\Sigma$ means computational complexity will be $\mathcal{O}(N^3)$~\cite{cormen2009introduction}. If observations are received at a rate of $\SI{10}{\hertz}$, then this approach will not be able to keep up with the rate of new observations. To speed up computation, each cell is assumed to be independent. This limits the filter from correlating the values of neighboring cells, but if the wildfire grows slowly, then this approximation can be effective. With this simplification, filtering is performed by $N$ one-dimensional EKFs, which greatly reduces the computational complexity to $\mathcal{O}(N)$. For each cell with initial mean $\mu_{0}=0$ and variance $\sigma_{0}=0.1$, the EKF update is defined as 
\begin{align}
\bar{\mu}_{t} &= \mu_{t-1} \\
\bar{\sigma}_{t} &= \sigma_{t-1} + q  \\
k &= \bar{\sigma}_{t}(\bar{\sigma}_{t} + r)^{-1} \\
\mu_{t} &= \bar{\mu}_{t} + k(y_t-\textbf{1}\{\bar{\mu}_{t}>0.5\})   \\
\sigma_{t} &= \bar{\sigma}_{t} - k\bar{\sigma}_{t}  
\end{align}
where $y_t$ is the sensor observation, $q$ is the process noise variance, and $r$ is the observation noise variance~\cite{thrun2005probabilistic}.

\begin{figure}
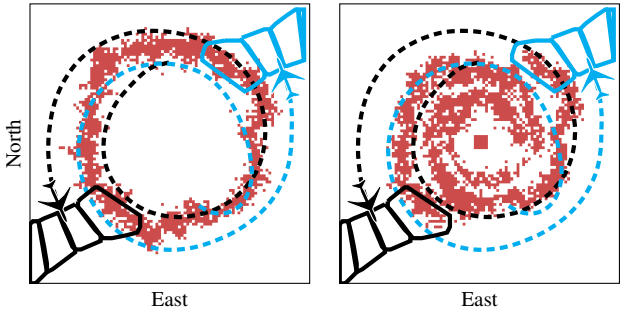

    \vspace{4pt}
	\include{EKF_Plot}
	\vspace{-15pt} 
	\caption{True wildfire (left) and EKF belief map (right)}
	\label{fig:EKF}
\end{figure}

While the EKF can filter and compile noisy images into a map, there are a few limitations. The EKF does not use any model of wildfire dynamics, so regions that are not observed are static until observed again. \Cref{fig:EKF} shows the final wildfire and aircraft configuration as well as the trajectories flown as the wildfire expands. The EKF mean is thresholded with $\hat{x}=1\{\mu>0.5\}$ to produce a wildfire belief map, which is accurate near the aircraft but less accurate at other points along or inside the fire front. In addition, the underlying wind parameters governing the wildfire growth remain uncertain. The next section discusses an approach that uses a particle filter to address these issues.

\section{Particle Filter}\label{sec:PF}
 An approach to filtering wildfire observations while estimating wind and wildfire growth is to use a particle filter where each particle represents a probabilistic wildfire model. As opposed to the discrete wildfire model presented in \cref{sec:wildfire}, this probabilistic model uses distributions over $B(s)$ and $F(s)$ because the true values are unknown. For each cell $s$, the probability that a cell is burning at time $t$ is $b_t(s)$ with fuel distribution $f^K_t(s)$, where $K\in[0,1,\dots,K_{\text{max}}]$ is the amount of fuel remaining. The wildfire probabilities are initialized with a small square of high probability wildfire and a uniform distribution over possible fuels, and then they are propagated according to 
 \begin{align}
 p(f^0_{t+1}(s)) &=  p(f^0_{t}(s)) + p(f^1_{t}(s))p(b_{t}(s))   \\
 p(f^K_{t+1}(s)) &=  p(f^K_{t}(s))(1-p(b_t(s))) +   \\
 &\phantom{{}=20} p(f^{K+1}_{t}(s))p(b_{t}(s)) \notag \\
 \rho(s) &=1-\prod_{s'} (1-P(s,s')p(b_t(s'))) \\
 p(b_t(s)) &= (1-p(f^0_{t}))[(1-p(b_t(s))\rho(s) + p(b_t(s))] 
 \end{align}
 where $P(s,s')$ is the probability that cell $s'$ will ignite cell $s$ as a function of wind, and $\rho(s)$ is the probability that cell $s$ will ignite from nearby cells. Therefore, the wildfire probabilities will grow according to non-dimensional wind coefficients. With each time step, the wildfire probabilities are updated and the wind is allowed to change according to random Gaussian noise, which allows the particle filter to adapt to changing winds. 
 
 Each observation is used to update the particle likelihood. For particle $x$, the posterior log-likelihood $\ell(x \mid o)$ is computed according to Bayes' Rule as 
 \begin{equation}
 \ell(x \mid o) \propto \sum_{i=1}^N \log p(o_i \mid x) + \log p(x)
 \end{equation}
 
 Initially all particles are assumed equally likely; however, as the log-likelihood is updated with new measurements, some particles will be more likely than others. After 20 wildfire steps, the particle filter is resampled with a weighted distribution proportional to the normalized particle likelihoods \cite{arulampalam2002tutorial}. Resampling removes unlikely particles and encourages the filter to explore more promising particles. After resampling, some particles will be duplicated, but the random changes added to the wind encourage the particle filter to model wildfires with different wind speeds. For this work 40 particles were used.
 
 In addition to updating particle likelihoods, the observations are used to update the particle wildfire probabilities. For observation $o_t$ correlated with ground cell $s$, the particle update follows Bayes' rule as
 \begin{align}
 p(b_t(s) \mid o_t=1) &\propto p(o_t=1 \mid b_t(s))p(b_t(s))  \\
 p(b_t(s) \mid o_t=0) &\propto p(o_t=0 \mid b_t(s))p(b_t(s)) 
 \end{align}
where $p(o_t=1 \mid b_t(s))$ and $p(o_t=0 \mid b_t(s))$ define the observation model. The probability of observing correctly was set to 0.8. 

\begin{figure*}
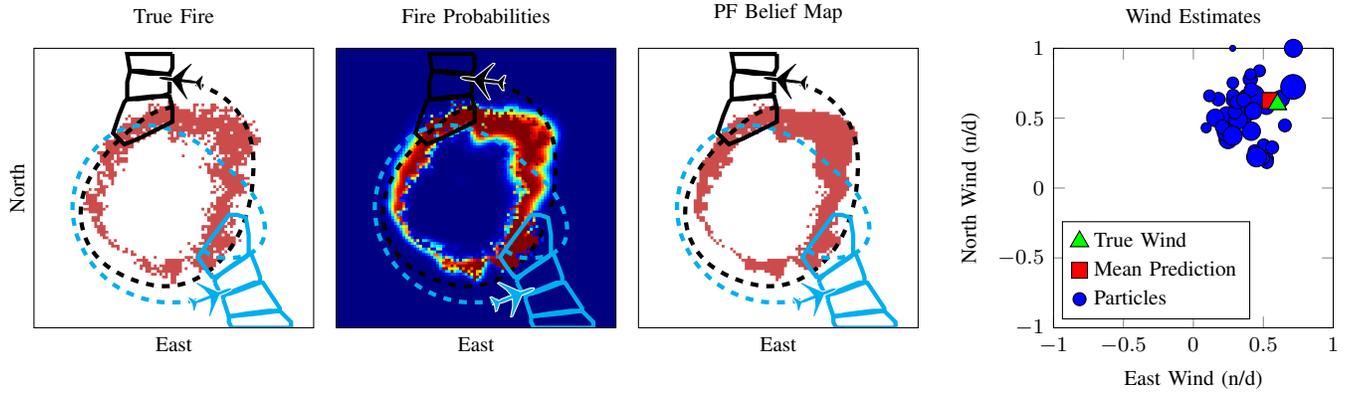

    \vspace{1pt}
	\include{PF_Plot}
	\vspace{-15pt} 
	\caption{Particle filter predictions for wildfire and wind}
	\label{fig:PF}
\end{figure*}

The particle filter's wildfire and wind predictions can be computed as the sum over particles weighted by normalized particle likelihood. \Cref{fig:PF} shows the state of a particle filter after a period where the aircraft have observed the wildfire. The fire probabilities can be thresholded as in the EKF approach to create a wildfire belief map that is accurate away from the aircraft, unlike the EKF belief map. Points inside the wildfire expansion have a low probability of burning because the fuel is likely depleted, and the particle filter can predict the wildfire expansion even without new observations. In addition, the particle filter accurately predicts the true wind parameters of the fire, as shown in the rightmost plot of \cref{fig:PF}, where marker size denotes particle probability.

While good observations will enhance the particle filter accuracy, observations in unimportant locations will lead to large differences between the predicted wildfire and true wildfire. Therefore, the aircraft need to fly strategic trajectories to observe as much new wildfire as possible to compute an accurate wildfire belief map.

\section{Baselines}
In the following experiments, three baseline methods for generating aircraft trajectories are compared with the DRL approaches. The first baseline chooses actions randomly, and the second baseline uses a simple heuristic method to fly near the wildfire to make observations. The third baseline uses a receding horizon approach where the wildfire belief map is used to plan a trajectory of $T$ steps, then $t<T$ steps are executed before re-planning. The trajectory is optimized using a coordinate descent approach with random restarts, which yields good trajectories without needing to evaluate all possible trajectories.

\section{Simulation Results} \label{sec:results}
Twenty random simulations were conducted for different DRL and baseline controllers. Each method was evaluated with different penalties for flying over the wildfire, and four metrics were extracted. The first metric counts the number of fire cells within \SI{40}{\meter} of the aircraft during the trajectory, while the second metric counts the number of fire cells observed during the trajectory. The third metric computes the Hamming distance between the thresholded belief map and true wildfire map, and the fourth metric computes the average error in predicted wind. Although the EKF approach does not approximate wind, the observations can be used with the particle filter even though the particle filter is not being used to guide the aircraft. 

\begin{figure}
	\begin{tikzpicture}[]
\begin{groupplot}[height = {4.46cm}, width = {4.46cm},group style={horizontal sep=1.4cm, vertical sep=1.4cm, group size=2 by 2}]
\nextgroupplot [ylabel = {Fire Cells Observed}, ymax = {3200}, xlabel = {Fire Cells Flown Over}, scaled ticks=base 10:-3,x tick scale label style={at={(xticklabel cs:0.97)},anchor=south west},y tick scale label style={at={(-0.08,1.02)},anchor=south west}, ymin = {0}]\addplot+ [mark = {*}, blue,mark options={blue}]coordinates {
	(23.599999999999998, 2044.35)
	(1230.7999999999997, 2377.1500000000005)
	(3368.1499999999996, 2745.5999999999995)
};
\addplot+ [mark = {square*}, red,mark options={red}]coordinates {
	(35.3, 2089.25)
	(1508.35, 2424.4)
	(4583.75, 2761.9000000000005)
};
\addplot+ [mark = {triangle*}, yellow!40!black,mark options={yellow!40!black}]coordinates {
	(139.5, 1358.05)
	(504.8000000000001, 1652.3499999999997)
	(1790.2999999999997, 2268.1)
	(8073.000000000001, 2415.8499999999995)
};
\addplot+ [mark = {star}, black,mark options={black}]coordinates {
	(1548.9499999999996, 1680.3500000000001)
	(5952.350000000001, 2662.4999999999995)
};
\addplot+[draw=none, mark = {diamond*}, cyan,mark options={cyan}] coordinates {
	(1869.2499999999998, 1237.3000000000002)
};
\nextgroupplot [ylabel = {Belief Map Error}, xlabel = {Fire Cells Flown Over}, scaled ticks=base 10:-3, x tick scale label style={at={(xticklabel cs:0.97)},anchor=south west},y tick scale label style={at={(-0.08,1.02)},anchor=south west},legend style={draw=none,at={(0.33,-0.5)},column sep=3pt,anchor=north}, ymin = {0}]\addplot+ [mark = {*}, blue,mark options={blue}]coordinates {
	(23.599999999999998, 1257.6)
	(1230.7999999999997, 933.7500000000001)
	(3368.1499999999996, 531.55)
};
\addlegendentry{EKF}
\addplot+ [mark = {square*}, red,mark options={red}]coordinates {
	(35.3, 393.44999999999993)
	(1508.35, 343.59999999999997)
	(4583.75, 238.35000000000002)
};
\addlegendentry{PF}
\addplot+ [mark = {triangle*}, yellow!40!black,mark options={yellow!40!black}]coordinates {
	(139.5, 665.45)
	(504.8000000000001, 547.2)
	(1790.2999999999997, 371.09999999999997)
	(8073.000000000001, 337.1)
};
\addlegendentry{Receding Horizon}
\addplot+ [mark = {star}, black,mark options={black}]coordinates {
	(1548.9499999999996, 555.15)
	(5952.350000000001, 291.35)
};
\addlegendentry{Heuristic}
\addplot+[draw=none, mark = {diamond*}, cyan,mark options={cyan}] coordinates {
	(1869.2499999999998, 859.0)
};
\addlegendentry{Random}
\addplot+ [mark = {o}, solid, green!50!black, mark options={thick,green!50!black}]coordinates {
	(23.599999999999998, 395.85)
	(1230.7999999999997, 372.75000000000006)
	(3368.1499999999996, 240.50000000000006)
};
\addlegendentry{EKF + PF}
\nextgroupplot [ylabel = {Wind Prediction Error}, xlabel = {Fire Cells Flown Over}, scaled ticks=base 10:-3,scaled y ticks=false, x tick scale label style={at={(xticklabel cs:0.97)},anchor=south west}, ymin = {0}]\addplot+ [mark = {square*}, red,mark options={red}]coordinates {
	(35.3, 0.18324790912692457)
	(1508.35, 0.1676320822982425)
	(4583.75, 0.14644735773816164)
};
\addplot+ [mark = {triangle*}, yellow!40!black,mark options={yellow!40!black}]coordinates {
	(139.5, 0.30650862014003805)
	(504.8000000000001, 0.3071210967600088)
	(1790.2999999999997, 0.2530325099015662)
	(8073.000000000001, 0.2171326551181197)
};
\addplot+ [mark = {star}, black,mark options={black}]coordinates {
	(1548.9499999999996, 0.24457105970913037)
	(5952.350000000001, 0.2525842879068416)
};
\addplot+[draw=none, mark = {diamond*}, cyan,mark options={cyan}] coordinates {
	(1869.2499999999998, 0.36821299919872463)
};
\addplot+ [mark = {o}, solid, green!50!black, mark options={thick,green!50!black}]coordinates {
	(23.599999999999998, 0.1923760360729977)
	(1230.7999999999997, 0.2122991490188133)
	(3368.1499999999996, 0.1851480574842027)
};
\end{groupplot}

\end{tikzpicture}
	\vspace{-10pt} 
	\caption{Simulation results for DRL and baseline methods}
	\label{fig:Results}
\end{figure}
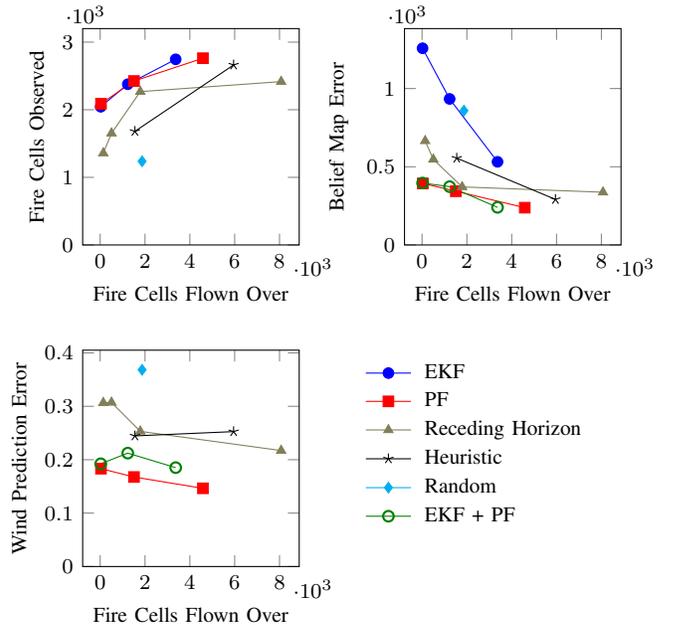

\Cref{fig:Results} shows the performance of the different methods as a function of fire cells flown over, which varies as each method is tuned with different penalties for flying over fire. Both DRL approaches dominate the baselines in all metrics. Although the EKF belief map does not represent the wildfire as well, applying the particle filter to the generated observations improves performance. Overall both the EKF and particle filter methods are effective in controlling the aircraft. These results suggest that the simpler EKF approach is as effective in guiding the aircraft as the particle filter.

\begin{figure}
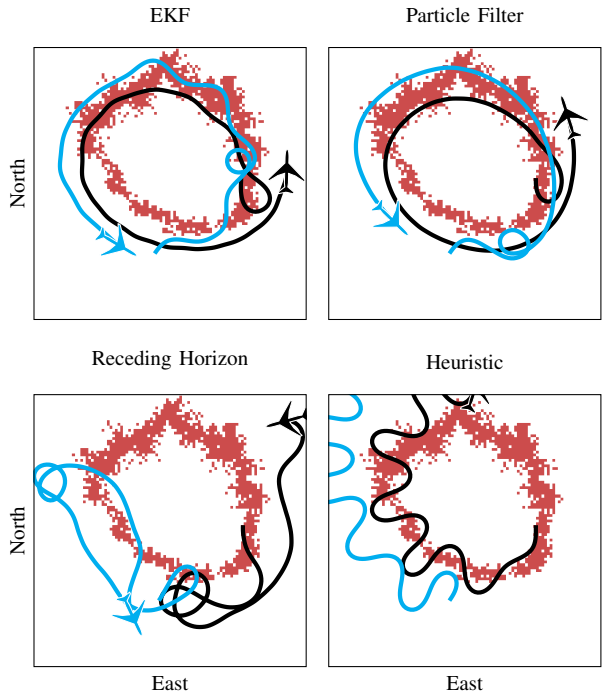

	\include{TrajComp}
	\vspace{-10pt} 
	\caption{Trajectories flown with penalty for flying over wildfire}
	\label{fig:Traj}
\end{figure}

\begin{figure}
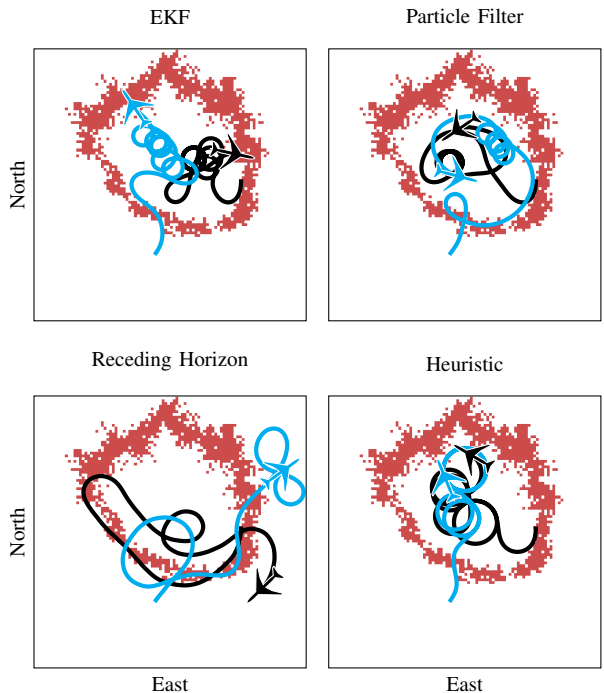

	\include{TrajComp_Fire}
	\vspace{-10pt} 
	\caption{Trajectories flown without penalty for flying over wildfire}
	\label{fig:TrajFire}
\end{figure}

To further study the behavior of the controllers, \cref{fig:Traj,fig:TrajFire} show the trajectories of the aircraft with different controllers. When flying over wildfire is discouraged, as in \cref{fig:Traj}, both DRL controllers circle around the wildfire at a safe distance. Furthermore, the cyan aircraft performs a stalling $360^\circ$ turn to gain separation from the other aircraft so observations are not redundant. This behavior illustrates that the aircraft can cooperate implicitly to monitor wildfire growth, while the baseline trajectories are uncoordinated and less effective. \Cref{fig:TrajFire} shows the trajectories flown when flying over wildfire is not penalized. The DRL approaches, along with the heuristic method, guide the aircraft to the interior of the fire and make tight turns, allowing the aircraft to view the entire front from one location.

\begin{figure}
    \vspace{4pt}
	\begin{tikzpicture}[]
\begin{groupplot}[height = {4.48cm}, width = {4.48cm}, group style={horizontal sep=1.4cm, vertical sep=1.4cm, group size=2 by 2}]
\nextgroupplot [ylabel = {Fire Cells Observed}, xlabel = {Fire Cells Flown Over}, scaled ticks=base 10:-3,x tick scale label style={at={(xticklabel cs:0.92)},anchor=south west},y tick scale label style={at={(-0.08,1.02)},anchor=south west}, ymin = {0}]\addplot+ [mark = {*}, blue,mark options={blue}]coordinates {
(23.599999999999998, 2044.35)
(1230.7999999999997, 2377.1500000000005)
(3368.1499999999996, 2745.5999999999995)
};
\addplot+ [mark = {square*}, cyan,mark options={cyan}]coordinates {
(35.3, 2089.25)
(1508.35, 2424.4)
(4583.75, 2761.9000000000005)
};
\addplot+ [mark = {*}, densely dotted,thick,green!40!black,mark options={green!40!black,solid}]coordinates {
(4.6, 1686.8)
(1413.8500000000004, 2202.5999999999995)
(3603.45, 2733.6)
};
\addplot+ [mark = {square*}, densely dotted,thick,green!80!black,mark options={green!80!black,solid}]coordinates {
(41.800000000000004, 2088.0)
(1380.7, 2417.0499999999997)
(4148.700000000001, 2762.1999999999994)
};
\addplot+ [mark = {*}, densely dashed,red,mark options={red,solid}]coordinates {
(356.3, 1686.8)
(3250.4499999999994, 2202.5999999999995)
(4257.45, 2733.6)
};
\addplot+ [mark = {square*}, densely dashed,red!50!white,mark options={red!50!white,solid}]coordinates {
(38.4, 2088.0)
(1317.7, 2417.0499999999997)
(4376.25, 2762.1999999999994)
};
\addplot+ [mark = {*}, dashed,brown!60!black,mark options={brown!60!black,solid}]coordinates {
(1511.7499999999998, 1061.8999999999999)
(3193.5, 1871.2999999999997)
(4137.4, 2164.75)
};
\addplot+ [mark = {square*}, dashed,brown!80!white,mark options={brown!80!white,solid}]coordinates {
(147.85, 1953.9999999999998)
(677.3500000000001, 2416.4)
(3288.000000000001, 2823.45)
};
\nextgroupplot [ylabel = {Belief Map Error}, xlabel = {Fire Cells Flown Over}, scaled ticks=base 10:-3,x tick scale label style={at={(xticklabel cs:0.92)},anchor=south west},y tick scale label style={at={(-0.08,1.02)},anchor=south west},legend columns=1,legend style={draw=none,at={(0.3,-0.49)},column sep=3pt,anchor=north}, ymin = {0}]\addplot+ [mark = {*}, blue,mark options={blue}]coordinates {
(23.599999999999998, 395.85)
(1230.7999999999997, 372.75000000000006)
(3368.1499999999996, 240.50000000000006)
};
\addlegendentry{EKF, 10\% Error}
\addplot+ [mark = {square*}, cyan,mark options={cyan}]coordinates {
(35.3, 393.44999999999993)
(1508.35, 343.59999999999997)
(4583.75, 238.35000000000002)
};
\addlegendentry{PF, 10\% Error}
\addplot+ [mark = {*}, densely dotted,thick,green!40!black,mark options={green!40!black,solid}]coordinates {
(4.6, 513.75)
(1413.8500000000004, 475.8499999999999)
(3603.45, 268.0)
};
\addlegendentry{EKF, 15\% Error}
\addplot+ [mark = {square*}, densely dotted,thick,green!80!black,mark options={green!80!black,solid}]coordinates {
(41.800000000000004, 386.70000000000005)
(1380.7, 364.2)
(4148.700000000001, 258.75)
};
\addlegendentry{PF, 15\% Error}
\addplot+ [mark = {*}, densely dashed,red,mark options={red,solid}]coordinates {
(356.3, 513.75)
(3250.4499999999994, 475.8499999999999)
(4257.45, 268.0)
};
\addlegendentry{EKF, 25\% Error}
\addplot+ [mark = {square*}, densely dashed,red!50!white,mark options={red!50!white,solid}]coordinates {
(38.4, 386.70000000000005)
(1317.7, 364.2)
(4376.25, 258.75)
};
\addlegendentry{PF, 25\% Error}
\addplot+ [mark = {*}, dashed,brown!60!black,mark options={brown!60!black,solid}]coordinates {
(1511.7499999999998, 2422.85)
(3193.5, 2446.95)
(4137.4, 2535.7)
};
\addlegendentry{EKF, 35\% Error}
\addplot+ [mark = {square*}, dashed,brown!80!white,mark options={brown!80!white,solid}]coordinates {
(147.85, 2798.0999999999995)
(677.3500000000001, 2532.1)
(3288.000000000001, 2045.1999999999996)
};
\addlegendentry{PF, 35\% Error}
\nextgroupplot [ylabel = {Wind Estimate Error}, xlabel = {Fire Cells Flown Over}, scaled ticks=base 10:-3,scaled y ticks=false,x tick scale label style={at={(xticklabel cs:0.92)},anchor=south west}, ymin = {0}]\addplot+ [mark = {*}, blue,mark options={blue}]coordinates {
(23.599999999999998, 0.1923760360729977)
(1230.7999999999997, 0.2122991490188133)
(3368.1499999999996, 0.1851480574842027)
};
\addplot+ [mark = {square*}, cyan,mark options={cyan}]coordinates {
(35.3, 0.18324790912692457)
(1508.35, 0.1676320822982425)
(4583.75, 0.14644735773816164)
};
\addplot+ [mark = {*}, densely dotted,thick,green!40!black,mark options={green!40!black,solid}]coordinates {
(4.6, 0.2557475632254919)
(1413.8500000000004, 0.3097254687994827)
(3603.45, 0.25189164623826066)
};
\addplot+ [mark = {square*}, densely dotted,thick,green!80!black,mark options={green!80!black,solid}]coordinates {
(41.800000000000004, 0.24014699204460388)
(1380.7, 0.22587345110109952)
(4148.700000000001, 0.1921433498961455)
};
\addplot+ [mark = {*}, densely dashed,red,mark options={red,solid}]coordinates {
(356.3, 0.2557475632254919)
(3250.4499999999994, 0.3097254687994827)
(4257.45, 0.25189164623826066)
};
\addplot+ [mark = {square*}, densely dashed,red!50!white,mark options={red!50!white,solid}]coordinates {
(38.4, 0.24014699204460388)
(1317.7, 0.22587345110109952)
(4376.25, 0.1921433498961455)
};
\addplot+ [mark = {*}, dashed,brown!60!black,mark options={brown!60!black,solid}]coordinates {
(1511.7499999999998, 0.9068350564639011)
(3193.5, 0.6507493757129348)
(4137.4, 0.6926463266538523)
};
\addplot+ [mark = {square*}, dashed ,brown!80!white,mark options={brown!80!white,solid}]coordinates {
(147.85, 0.7390930401627424)
(677.3500000000001, 0.42182400918302776)
(3288.000000000001, 0.44461944289985567)
};
\end{groupplot}

\end{tikzpicture}
	\vspace{-10pt} 
	\caption{Simulation results for DRL methods with observation errors}
	\label{fig:ErrorRateResults}
\end{figure}
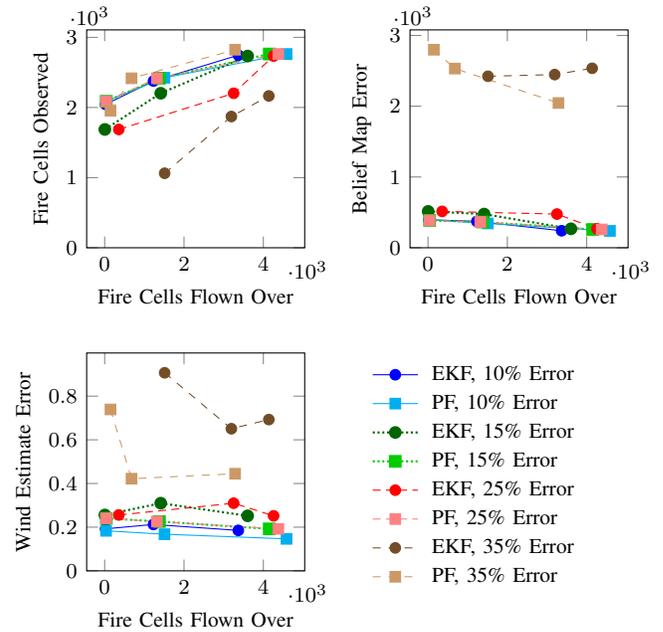

Increasing observation error shows one advantage of the particle filter over the EKF for controlling aircraft. As shown in \cref{fig:ErrorRateResults}, increasing the observation error level from 10\% decreases the number of fire cells observed when using the EKF but not when using the particle filter. The particle filter is less easily fooled by noisy observations because it uses a prior over wildfire locations, although the generated belief map and wind estimates degrade as the error level reaches 35\%. This result demonstrates that the particle filter controller is more robust to observation error.

\section{Conclusions}
This work presented deep reinforcement learning (DRL) controllers for wildfire surveillance using noisy camera observations. Two methods, an extended Kalman filter (EKF) and particle filter, were implemented to filter the images into a belief map, which were used as the input to the DRL controller. Simulations were conducted with different penalties for flying over the wildfire, and both DRL methods were shown to effectively guide the aircraft to survey the wildfire and outperformed baseline methods. The particle filter was able to accurately predict wildfire growth and wind parameters, which are informative indicators about the state of the wildfire. In addition, the particle filter was shown to be more robust to observation error than the EKF because it uses a prior over the wildfire locations. Animations and code can be found at github.com/sisl/UAV\_Wildfire\_Monitoring.

\section*{Acknowledgments}
The authors wish to thank Jeremy Morton for his helpful feedback. This material is based upon work supported by the National Science Foundation Graduate Research Fellowship under Grant No. DGE-1656518. Any opinion, findings, and conclusions or recommendations expressed in this material are those of the authors and do not necessarily reflect the views of the National Science Foundation.

\printbibliography

\end{document}